\documentclass[sigconf,authorversion]{acmart}

\usepackage{balance}

\def\BibTeX{{\rm B\kern-.05em{\sc i\kern-.025em b}\kern-.08emT\kern-.1667em\lower.7ex\hbox{E}\kern-.125emX}}
    
\setcopyright{acmcopyright}

\copyrightyear{2019} 
\acmYear{2019} 
\acmConference[CIKM '19]{The 28th ACM International Conference on Information and Knowledge Management}{November 3--7, 2019}{Beijing, China}
\acmBooktitle{The 28th ACM International Conference on Information and Knowledge Management (CIKM '19), November 3--7, 2019, Beijing, China}
\acmPrice{15.00}
\acmDOI{10.1145/3357384.3358117}
\acmISBN{978-1-4503-6976-3/19/11}
\acmSubmissionID{sp1174}

\usepackage{latexsym}
\usepackage{url}
\usepackage{hyperref}
\usepackage{tabularx}
\usepackage{amsmath}
\usepackage{booktabs}
\usepackage{xspace}
\usepackage{amssymb}
\usepackage{amsmath}
\usepackage{multirow}
\usepackage{CJKutf8}
\usepackage{arydshln}

\newcommand{\ie}{i.e.,\xspace}
\newcommand{\eg}{e.g.,\xspace}
\newcommand{\paratitle}[1]{\noindent\textbf{#1}}

\newcommand{\secbest}[1]{\textcolor{blue}{\textit{\textbf{#1}}}}
\newcommand{\baby}{\textsc{ME-CNER}\xspace}

\begin{document}

\fancyhead{}

\title{Exploiting Multiple Embeddings for Chinese Named Entity Recognition}

\author{Canwen Xu}
\authornote{These two authors contribute equally to this work.}
\orcid{0000-0002-1552-999X}
\affiliation{
  \institution{Wuhan University}
  \department{School of Computer Science}
}
\email{xucanwen@whu.edu.cn}

\author{Feiyang Wang}
\authornotemark[1]
\affiliation{
  \institution{Wuhan University}
  \department{School of Computer Science}
}
\email{wangfeiyang@whu.edu.cn}

\author{Jialong Han}
\affiliation{
  \institution{Tencent AI Lab}
}
\email{jialonghan@gmail.com}

\author{Chenliang Li}
\authornote{Chenliang Li is the corresponding author.}
\affiliation{
  \institution{Wuhan University}
  \department{School of Cyber Science and Engineering}
}
\email{cllee@whu.edu.cn}

%
\renewcommand{\shortauthors}{C. Xu et al.}

%
\begin{abstract}
Identifying the named entities mentioned in text would enrich many semantic applications at the downstream level. However, due to the predominant usage of colloquial language in microblogs, the named entity recognition (NER) in Chinese microblogs experience significant performance deterioration, compared with performing NER in formal Chinese corpus. In this paper, we propose a simple yet effective neural framework to derive the character-level embeddings for NER in Chinese text, named \baby. A character embedding is derived with rich semantic information harnessed at multiple granularities, ranging from radical, character to word levels. The experimental results demonstrate that the proposed approach achieves a large performance improvement on Weibo dataset and comparable performance on MSRA news dataset with lower computational cost against the existing state-of-the-art alternatives.
\end{abstract}

%
%


%

\maketitle

\section{Introduction}
\label{sec:intro}
As a fundamental task in NLP, Named Entity Recognition (NER) has been studied extensively in the past decade. Formally, NER is a task that locates the mentions of named entities and classifies their types like person, organization, location in unstructured text. Nowadays, the sources of text are various. Except for formal text like news, user-generated content like micro-blogging has played a central role in people's digital life. 
Performing NER over this real-time information tank of high volume would breed many downstream semantic applications. 

\begin{CJK*}{UTF8}{gbsn}
Compared with the English language, NER on Chinese text faces more challenges. 
It is known that Asian languages like Chinese are naturally logographic. There are no conventional linguistic features (like capitalization) available for NER. Moreover, Chinese characters and words hold complex relations. Given no delimiter between characters, a word can be comprised of a single character or multi-character (\ie n-char). Also, the meaning of a Chinese word can be inferred by its constituent characters. For instance, the meaning of ``自信'' (confidence) can be derived by ``自'' (self) and ``信'' (belief). Furthermore, Chinese characters often have hieroglyphic forms. The same radical often indicates similar semantics. Besides, to build a new large NER dataset reflecting new words and memes over Chinese text is often expensive and labor-intensive. The available dataset is often dated or limited in size, which calls for a low-cost mechanism to harness the background knowledge from large unlabeled resources.
\end{CJK*}

\begin{figure}
\includegraphics[width=7.7cm]{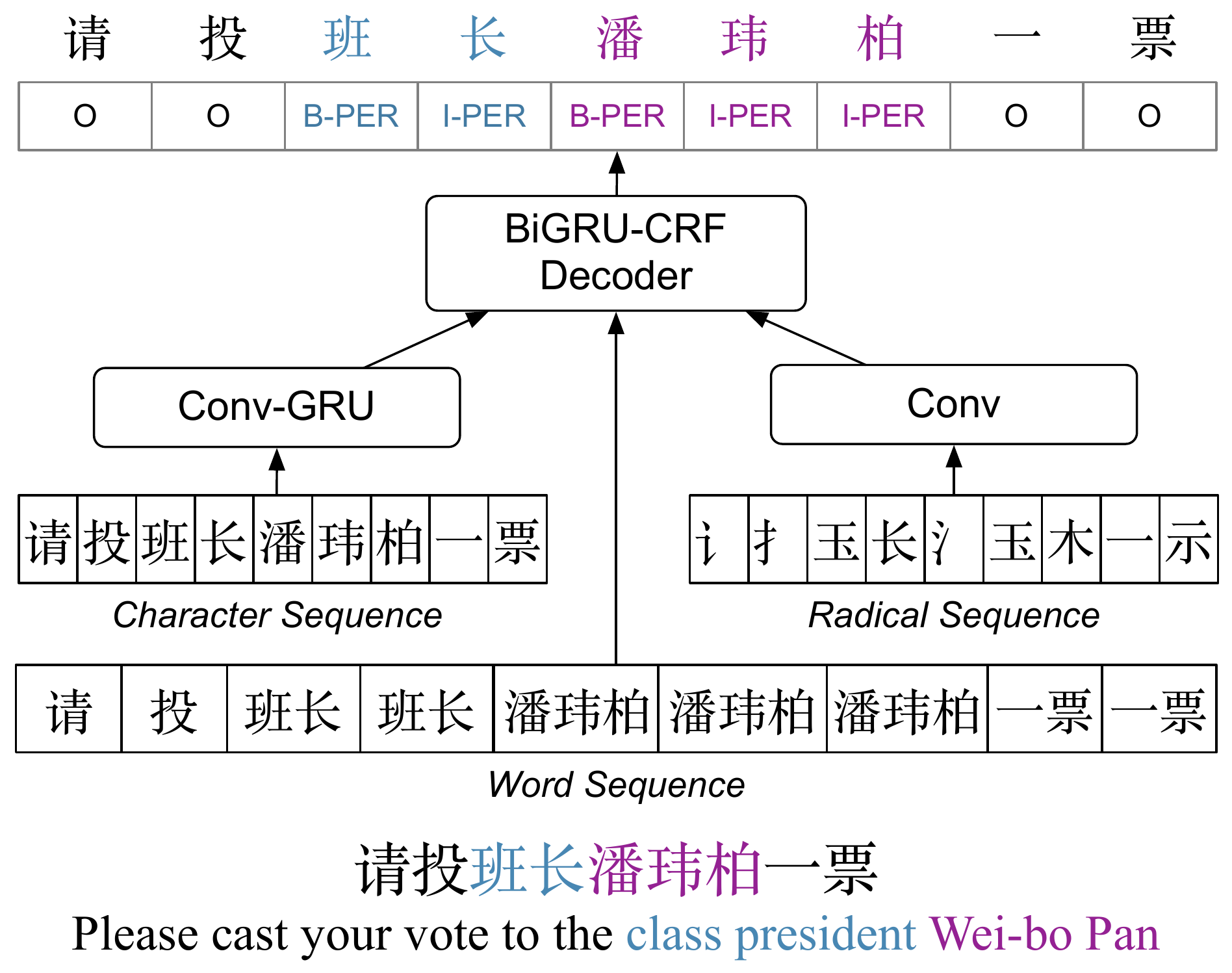}
\caption{A live example for the proposed architecture. ``班长'' is a nominal mention, while ``潘玮柏'' is a named entity.}
\label{datastream} 
\end{figure}

To this end, in this paper, we propose a simple yet effective neural framework for NER in Chinese text, named \baby. Specifically, we derive a character representation based on multiple embeddings in different granularities, ranging from radical, character to word levels. We utilize a convolutional operation to derive the character representation based on the local radical-level context information. Also, a Conv-GRU network is designed to capture semantic representation for a character based on its local context and long-range dependence. To harness the background knowledge, we choose to initialize the embeddings at character- and word- levels with pre-trained embeddings over a large external corpus, and fine-tune them in a task-driven fashion. As illustrated in Figure \ref{datastream}, these character representations in different perspectives are concatenated and fed into a BiGRU-CRF (Bidirectional Gated Recurrent Unit - Conditional Random Field) tagger for sequence labeling. The experimental results over a standard benchmark Weibo (Chinese Microblog) dataset show that more than $7.6\%$ relative performance improvement is obtained against the existing state-of-the-art solutions. Also, we provide results on a large and formal dataset (contrary to Weibo), MSRA News, to confirm the effectiveness of our proposed method. To summarize, we make the following contributions:
(1) We propose a multiple-embedding-driven neural framework for Chinese NER. Various semantic information is well exploited in different granularities. (2) Our method significantly outperforms the previous state-of-the-art with a large margin on Weibo dataset and achieves comparable performance on MSRA News dataset with a lower computational cost. Either Conv-GRU or radical embedding is revealed to be beneficial for the task. Our code is available at \url{https://github.com/WHUIR/ME-CNER}.

\begin{CJK*}{UTF8}{gbsn}
\end{CJK*}

\section{Related Work}
Many Deep Learning techniques have been proposed for NER task.
\citet{corr15:Huang} proposed a BiLSTM~(Bidirectional Long Short-Term Memory)-CRF~(Conditional Random Field) network as NER tagger. \citet{naacl16:Lample} used a BiLSTM to extract word representations over character-level embeddings. \citet{acl16:Ma} used a CNN~(Convolutional Neural Network)-BiLSTM-CRF network to utilize both word and character level representations.

Chinese NER is more challenging compared to English. Many sophisticated methods have been proposed in this line.  \citet{ccl17:Wang} proposed a gated CNN model. \citet{cikm17:e} utilized character and word mixed embeddings. \citet{acl18:Zhang} proposed Lattice LSTM model. As to NER in social media, \citet{emnlp15:peng} used the positional character embedding and jointly trained it with word embedding. \citet{emnlp18:Cao} used adversarial transfer learning. \citet{acl16:Peng} utilized word segmentation neural network to facilitate NER. \citet{eacl17:He} proposed an integrated method to train on
both F-score and label accuracy. Note that either character and word level embeddings are exploited in these works. However, \baby is the first work to jointly consider radical, character and word level embeddings for Chinese NER.

\section{Model Structure and Multiple Embeddings}
\begin{figure}
\includegraphics[width=7.7cm]{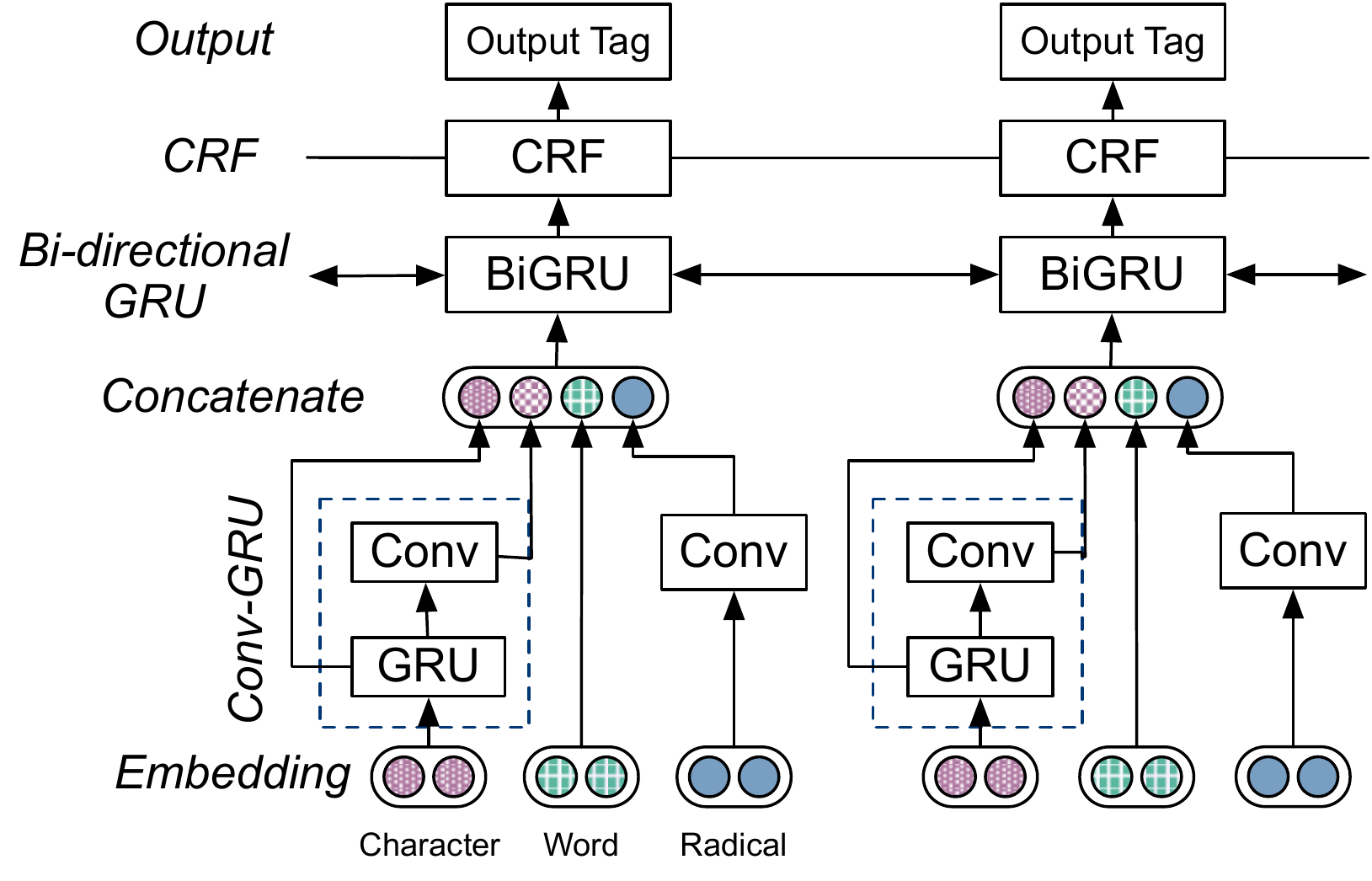}
\caption{The overall neural architecture of \baby.}
\label{structure}
\end{figure}

In this section, we introduce the multiple embeddings from finer to coarse level of granularity. The overall structure of \baby is shown in Figure \ref{structure}.

\subsection{Radical Embedding}
\begin{CJK*}{UTF8}{gbsn}
Chinese characters are hieroglyphic in nature. The same parts (\ie radical) in different characters often share the same meaning~\cite{linguistic:ho}.
The motivation for the use of radicals for NER is intuitive yet cultural. Chinese speakers have a strong inclination to use ``good characters'' when naming their businesses and children. For example, characters with radicals like ``疒(disease)'' and ``尸(death)'' never appear in a name of a person or a business. On the contrary, characters with radicals including ``钅(metal)'', ``木(wood)'', ``氵(water)'' and ``火(fire)'' are very common in a person's name to meet the Wu-Xing theory (a Chinese folk belief).
\end{CJK*}

We use a CNN network to extract local context features of the radical sequence. The use of radicals also enables us to better infer the semantics of the characters that only appear in the test set but not in the training set. That is, we can teach the model to generalize well for rare characters correctly.

\subsection{Character Embedding}
\label{subsec:ce}
Characters are the elementary units in Chinese. Note that each Chinese character has its own meaning and can compose a word itself. Also, the meaning of a Chinese word can be inferred by considering its constituent characters. Thus, it is critical to exploit the rich semantics contained by characters.

Previous studies use character-level CNN model for NER tasks~\cite{ccl17:Wang,acl16:Ma}. However, CNN emphasizes the local n-char features within a specific window and is not capable of capturing long-range dependence. To deal with this problem, we propose Convolutional Gated Recurrent Unit (Conv-GRU) Network which is composed of a GRU layer and a convolution layer. First, the character embedding $\mathbf{c}_i$ is fed into the GRU layer.

\begin{equation}
     \mathbf{x}_i = \textrm{GRU}(\mathbf{c}_1, ..., \mathbf{c}_i)
 \end{equation}
 Then the output $\mathbf{X}=[\mathbf{x}_1,...,\mathbf{x}_l]$ is fed into a convolutional layer padded to the same length as the input, where $l$ is the length of the microblog.
 \begin{equation}
     \mathbf{Y} = \textrm{Conv}(X)
 \end{equation}
 Finally, the output of the convolutional layer is concatenated with the output of the GRU layer to form the final representation for each character.
 \begin{equation}
     \mathbf{z}_i = \mathbf{x}_i \oplus \mathbf{y}_i
 \end{equation}
 
In this way, we can combine the semantic knowledge from both local context and long-term dependency together. 

\begin{table}[t!]
  \centering
  \begin{tabular}{l|c|rrr}
    \toprule
    Dataset & Type & Train & Valid & Test \\
    \midrule
    \multirow{2}{*}{Weibo} & Sentence & 1.4k & 0.27k & 0.27k \\ 
    & Character & 73.8k & 14.5k & 14.8k \\
    \midrule
    \multirow{2}{*}{MSRA} & Sentence & 46.4k & - & 4.4k \\
    & Character & 2169.9k & - & 172.6k \\
    \bottomrule
  \end{tabular}
  \caption{Statistics of the two datasets used in our experiments.}
  \label{tab:stat}
\end{table}

\subsection{Word Embedding}
\begin{CJK*}{UTF8}{gbsn}
As a high-level representation, word embedding is the most common way to exploit semantics in NER. However, a Chinese word would be a multi-characters. As shown in Figure~\ref{datastream}, to align the segmented words and the characters, we duplicate the word embedding for its constituent characters. For example, for the example demonstrated in Figure~\ref{datastream}, both component character ``班'' (class) and ``长'' (president) are aligned with a shared word embedding of ``班长(class president)''. If a word is not in the vocabulary of the word embedding, we initialize its embedding with random values.
\end{CJK*}

\subsection{BiGRU-CRF Tagger}
We concatenate the character embeddings in radical, character and word levels to form the final character representation. Then, we utilize the BiRNN-CRF~\cite{corr15:Huang} tagger to label each sentence. BiGRU-CRF is composed of forward and backward GRU layers and a CRF layer on top of the former. Here, we initialize the word embeddings and character embeddings with pre-trained embedding provided by Tencent AI Lab~\cite{naacl18:song} (\url{https://ai.tencent.com/ailab/nlp/embedding.html}).

\section{Experiments}
\label{sec:exp}

In this section, we evaluate the proposed \baby against the existing state-of-the-art models. Then we carry out ablation studies to verify each design choice in our model.

\subsection{Experimental Settings}
\paratitle{Dataset.} We utilize a standard Weibo NER dataset provided by~\cite{emnlp15:peng}. It consists of both named entities and nominal mentions. Also, we conduct experiments on a formal text dataset, MSRA News dataset~\cite{acl06:Levow}, which only includes named entities. The statistics of the two datasets we use is shown in Table \ref{tab:stat}.

Following the related works~\cite{emnlp15:peng,cikm17:e}, we perform Chinese word segmentation with Jieba toolkit~(\url{https://github.com/fxsjy/jieba}). The character-level radicals are extracted according to a look-up table provided in~\cite{emnlp17:Yu}.

\paratitle{Parameter Setting and Evaluation Metrics.}
We set the size of the kernel of the convolutional layers to $3$ for both radical and character level embeddings. In the Conv-GRU framework, the dimension of GRU is set to $150$. The embedding size is fixed to be $200$. We add dropout with a keep probability of $0.8$. We run each experiment five times, then report average precision, recall and $F_1$ scores. For BiGRU tagger, the BIO scheme~\cite{aclvlc95:ramshaw} is utilized.

\subsection{Performance Comparison}
Table~\ref{result} summarizes the results of different models. When evaluated on Weibo dataset, \baby improves $22.13$ on precision (Pr), $2.14$ on recall (Re), and $10.14$ in terms of $F_1$, against the best baseline (\ie Lattice~\cite{acl18:Zhang}).We would like to attribute the improvement to the following enhancements. First, our use of multiple embeddings could encode semantic information towards NER from different perspectives. Second, the Conv-GRU networks work well in exploiting character-level information, which is extremely critical in Chinese NER task. Finally, the pre-trained embeddings are learned over large-scale news, crawled web pages and online novels, which provide further background knowledge. Note that all baselines also utilize the word and character embeddings pre-trained over a large external corpus. To examine this factor further, we train Lattice with Tencent embeddings. The results show that Tencent embeddings could further enhance the NER performance for Lattice. However, our proposed \baby still outperforms it by $4.84$ in terms of $F_1$.

On MSRA news dataset, our model performs slightly poorer than Lattice but outperforms all other baselines. To analyze, firstly, on this much larger dataset (see Table \ref{tab:stat}) the relatively simple architecture of our proposed model may suffer from a lack of learning capability. Also, the advantage of pre-trained embeddings fades with more supervised data. Moreover, on a small and noisy dataset, radicals can help generalize (to be detailed in Section \ref{sec:ablation}) but such advantage again vanishes in a formal and much larger dataset. However, we would like to highlight the lower computational cost ($< 0.5\times$ training time of Lattice) and better parallelism of our model brought by GRU and CNN, compared to Lattice, a complex modified LSTM network, which cannot implement batch training so far.

\begin{table*}[t!]
  \centering
  \begin{tabular}{l|ccc|ccc}
    \toprule
    \multirow{2}{*}{Method} &
    \multicolumn{3}{c|}{Weibo} &
    \multicolumn{3}{c}{MSRA News}\\
     & Pr & Re & F1 & Pr & Re & F1\\
    \midrule
    CPM (\citet{acl06:Chen}) & - & - & - & 91.22 & 81.71 & 86.20\\
    Joint(cp) (\citet{emnlp15:peng}) & 63.84 & 29.45 & 40.38 & - & - & -\\
    JTLE (\citet{acl16:Peng}) & 61.64 & 38.55 & 47.43 & - & - & -\\
    CiLin (\citet{lrec16:Lu}) & - & - & - & - & - & 87.94\\ 
    BLSTM-CRF+pretrain (\citet{nlpcc16:Dong}) & - & - & - & 91.28 & 90.62 & 90.95\\
    CWME (\citet{cikm17:e}) & \secbest{65.29} & 39.71 & 49.47 & - & - & -\\
    F-Score Driven (\citet{eacl17:He}) & - & - & 54.82 & - & - & -\\
    ATL (\citet{emnlp18:Cao}) & 55.72 & 50.68 & 53.08 & - & - & -\\
    \hdashline
    Lattice (\citet{acl18:Zhang}) & 53.04 & \secbest{62.25} & \secbest{58.79} & \textbf{93.57} & \textbf{92.79} & \textbf{93.18}\\
    Lattice with Tencent Embedding & 69.97 & 59.11 & 64.09 & 92.84 & 91.68 & 92.26\\
    \hline
    \baby \textit{(ours)}& \textbf{75.17} & \textbf{64.39} & \textbf{68.93} & \secbest{91.57} & \secbest{91.33} & \secbest{91.45}\\
    \hdashline
    - radical & 73.00 & 63.46 & 67.90 & 91.52 & 90.52 & 91.02\\
    - radical, Conv-GRU $\rightarrow$ CNN & 71.15 & 62.36 & 66.46 & 90.73 & 89.37 & 90.05\\
    - radical, Conv-GRU $\rightarrow$ BiLSTM & 71.12 & 60.69 & 65.78 & 91.43 & 90.04 & 90.73\\
    \bottomrule
  \end{tabular}
  \caption{Experiment results and ablation study on Weibo NER and MSRA datasets. The best and second-best performance is indicated with black boldface and blue italics respectively. For the ablation study, we first remove radical embedding (marked as ``-radical''), then replace Conv-GRU with CNN or BiLSTM (marked as ``Conv-GRU$\rightarrow$CNN'' and ``Conv-GRU$\rightarrow$BiLSTM'' respectively).}
  \label{result}
\end{table*}

\subsection{Ablation Study}
\label{sec:ablation}
We further conduct ablation study for \baby by removing radical embeddings and replacing Conv-GRU network with either a CNN or a BiLSTM network. Table~\ref{result} demonstrates the performance comparison.

\paratitle{Radical Embedding.} It is obvious that \baby experiences performance degradation to some extent without using radical level embeddings. As mentioned previously, radicals would abstract the meaning of similar words. We analyze its effect for nominal mentions and named entities by showing actual cases. 

\begin{CJK*}{UTF8}{gbsn}

For nominal mentions, the word consisting of a single character ``娘 (mother)'', means ``mother'' in older ages, and is gradually replaced by another character ``妈'' (mother) in modern days. This character is especially rare in social media, including our Weibo corpus. However, with the help of radical embeddings, our model can correctly label this character as \textit{PERSON}, because it shares the same radical ``女'' (woman).

For named entities, since ``高 (Gao)'' is a common Chinese surname, without radical embedding, ``高跟鞋 (high heels)'' is recognized as ``PER''. However, radical ``足 (foot)'' and ``革 (leather)'' seldom appear in a person's name, which helps our model correctly label the word as ``O'' (\ie not an entity). 

\paratitle{Conv-GRU.} Another observation is that either CNN or BiLSTM delivers significant poorer performance. This suggests that the Conv-GRU network is indeed more effective to derive semantic knowledge for NER. For example, in phrase ``嗓门小'' (the voice is soft), the CNN model labels the clause as [B-PER, I-PER, I-PER], because ``小-'' (little) is often used at the beginning of a nickname (\eg ``小李'' (Lil-Li)). However, in this clause, the phrase is not likely a name with ``小'' (little) placing at the rear. With the temporal information, Conv-GRU is able to label it correctly. 

\end{CJK*}

\section{Conclusion}
In this paper, we exploit multiple embeddings for Chinese NER. Also, we propose Conv-GRU to better exploit character embedding. Our proposed \baby gains a prominent improvement against state-of-the-art alternatives when testing on a real-world microblog dataset and achieves comparable performance on a large formal news dataset with a lower computational cost. Our work shows the effectiveness of exploiting multiple embeddings together in different granularities.

%
\begin{acks}
We would like to thank the anonymous reviewers for their great insights.
This research was supported by National Natural Science Foundation of China (No.~61872278). Chenliang Li is the corresponding author.
\end{acks}
%
\bibliographystyle{ACM-Reference-Format}
\bibliography{sample-base}

\end{document}